\documentclass[letterpaper, 10 pt, conference]{ieeeconf}  

\IEEEoverridecommandlockouts                              

\usepackage[linesnumbered,lined,ruled,commentsnumbered]{algorithm2e}
\usepackage{algpseudocode} 
\usepackage{graphics} 
\usepackage{epsfig}
\usepackage{multirow}
\usepackage{numprint} 
\usepackage{caption}
\usepackage{amsmath,amssymb,amsfonts}

\usepackage{tikz,xcolor,hyperref}
\definecolor{lime}{HTML}{A6CE39}
\DeclareRobustCommand{\orcidicon}{
	\begin{tikzpicture}
	\draw[lime, fill=lime] (0,0) 
	circle [radius=0.16] 
	node[white] {{\fontfamily{qag}\selectfont \tiny ID}};
	\draw[white, fill=white] (-0.0625,0.095) 
	circle [radius=0.007];
	\end{tikzpicture}
	\hspace{-2mm}
}
\foreach \x in {A, ..., Z}{\expandafter\xdef\csname orcid\x\endcsname{\noexpand\href{https://orcid.org/\csname orcidauthor\x\endcsname}
			{\noexpand\orcidicon}}
}

\title{\LARGE \bf
Evaluating Pedestrian Risks in Shared Spaces Through Autonomous Vehicle Experiments on a Fixed Track
}

\author{Enrico Del Re\orcidE{}, \emph{Graduate Student Member, IEEE}, Novel Certad\orcidN{}, \emph{Graduate Student Member, IEEE} \\ Joshua Varughese \orcidJ{}, \emph{Member, IEEE}, Cristina Olaverri-Monreal\orcidC{} \emph{Senior Member, IEEE}
\thanks{
        Johannes Kepler University Linz, Austria; Department Intelligent Transport Systems \texttt{enrico.del\_re@jku.at}
        }
    }

\begin{document}

\maketitle
\thispagestyle{empty}
\pagestyle{empty}

\begin{abstract}
The majority of research on safety in autonomous vehicles has been conducted in structured and controlled environments. However, there is a scarcity of research on safety in unregulated pedestrian areas, especially when interacting with public transport vehicles like trams. This study investigates pedestrian responses to an alert system in this context by replicating this real-world scenario in an environment using an autonomous vehicle. The results show that safety measures from other contexts can be adapted to shared spaces with trams, where fixed tracks heighten risks in unregulated crossings.
\end{abstract}

\section{INTRODUCTION}
\label{sec:introduction}

Traffic safety involving autonomous vehicles (AVs) has primarily been studied in conventional road environments, where interactions follow predictable traffic rules. However, as shown \cite{Morales_2019} and \cite{olaverri2020promoting}, when informal rules apply, such as in shared spaces, approaches that are effective in traditional environments to promote vulnerable road user (VRU) safety, like the use of visual communication cues, may not necessarily be applicable.
This unique dynamic requires careful consideration of safety standards and interaction protocols. In this context, \cite{tram_pedestrian_accident_data} found that a significant proportion of tram-pedestrian accidents occur near tram stops, where trams move at low speeds and pedestrians may cross without designated crosswalks.
In addition,  public transport vehicles like trams, operating in pedestrian zones, are generally afforded priority over pedestrians.

A common tool for analyzing the severity of traffic conflicts is Surrogate Safety Measures (SSMs), which correlate kinematic indicators with accident risk \cite{SSM}. SSMs are widely utilized by Advanced Driver Assistance Systems (ADAS) to safely navigate traffic and are incorporated into standards for AV functions. For instance, UN Regulation No. 157 \cite{UN157} outlines the conditions under which lane-changing assistants can be deployed using SSMs as part of the assessment framework.

Public transport vehicles with high levels of automation, such as self-driving subways or monorails, typically rely on distinct safety assessments. This is largely because these vehicles operate in controlled environments, such as closed tracks or elevated pathways, rather than in mixed traffic. 
Fixed-track vehicles like self-driving subways and trams face higher collision risks as they cannot avoid obstacles through maneuverability and have long braking distances, making emergency stops more dangerous.
As a result, there is a lack of understanding regarding how, or if, SSMs could be effectively applied to rail-bound systems interacting with pedestrians, especially for heavy public transportation such as trams where maneuverability constraints heighten safety concerns.

To address this gap, we conducted an experiment to examine pedestrian behavior in response to an AV operating on a fixed track in a pedestrian zone. The study focused on analyzing how pedestrians reacted to warnings about obstructing the vehicle’s path, evaluating the safety risks using Surrogate Safety Measures (SSMs) from a traffic accident perspective.
We define for the distributions of the SSMs the following:
\begin{itemize}
    \item Null Hypothesis ($H_0$): The mean values of the distributions for warning and no-warning (baseline condition) interactions are equal, indicating that warnings have no effect on safety.
    \item Alternative Hypothesis ($H_1$): The mean of the distribution for warning interactions is significantly higher than that of no-warning interactions, suggesting that warnings lead to safer outcomes.

\end{itemize}

The findings offer valuable insights for improving the safety of interactions between pedestrians and trams in shared spaces.

We introduce related research in section \ref{sec:RelatedWork}, highlighting the gap this work aims to address. In section \ref{sec:experiment} we detail the experiment conducted, with the methodology to analyze the recordings in section \ref{sec:Methodology}. The results of the analysis are presented in section \ref{sec:results}, with future work and conclusions completing the study in section \ref{sec:Conclusion}.

\section{Related Work}
\label{sec:RelatedWork}

While the safety assessment of tram-pedestrian interactions has received little to no attention in existing literature, there is a substantial body of work focused on vehicle-pedestrian interactions.

Several studies have examined pedestrian crossing scenarios involving AVs. For example, the authors \cite{Morales_2020} developed a framework to estimate pedestrian crossing intentions, aiming to reduce the uncertainty caused by the lack of eye contact between road users. This framework was tested in a real vehicle deployment.
Pedestrian crossing intentions at unsignalized crossings were also analyzed and predicted in \cite{Zhang_2023} and \cite{Moreno_2023}, with the authors in \cite{Zhang_2023} utilizing a neural network (multilayer perceptron) for predictions and the authors in \cite{Moreno_2023} utilizing a random forest classifier.

The authors of \cite{Miguel_2019} analyzed the impact of different graphical user interfaces (GUI) mounted on the AV to communicate with pedestrians to improve the perception of road safety. Several GUIs were tested and perceived positively.

In \cite{Hussein_2016} the authors proposed a collision prediction algorithm based on Pedestrian to Vehicle (P2V) and Vehicle to Pedestrian (V2P) communication. The algorithm is based on the distance and time to the collision point for both pedestrian and vehicle.

SSMs were used in the analysis of pedestrian-vehicle interactions by \cite{Intercultu_TTC} and \cite{Predicted_PET}. In \cite{Intercultu_TTC} the authors utilize Time-to-Collision (TTC), an SSM also utilized in this study, to compare vehicle-pedestrian conflicts across different cultural contexts. They also evaluate the safety metric in non-compliance scenarios, such as pedestrians crossing against a red light.
Similarly, \cite{Predicted_PET} explores the use of artificial intelligence models to predict potentially hazardous Post-Encroachment Time (PET) values, an SSM utilized in this study, in pedestrian-vehicle interactions. The study demonstrates how such predictions enable earlier initiation of safety maneuvers, contributing to improved safety outcomes.

An extensive review of research on Pedestrian Collision Avoidance (PAC) systems for AVs is provided by \cite{PAC_review}. The authors notably highlight the limited use of small-scale robotic platforms for testing PAC systems. In contrast, the experiment detailed in Section \ref{sec:experiment} leverages such a platform, which is advantageous as implementing the same study with larger AVs, such as cars, would raise significant safety concerns.

Interactions between fixed track AVs and pedestrians are mainly found in publications focused on accident statistics \cite{tram_pedestrian_accident_data} or passive safety systems \cite{tram_passive}. To the best of the authors' knowledge, no existing publications examine the trajectories and safety metrics of interactions between pedestrians and fixed-track AVs. This study seeks to fill that research gap with experimental data and detailed analysis.

\section{Experimental setup}
\label{sec:experiment}

\begin{figure}
    \centering
    \includegraphics[scale=0.25]{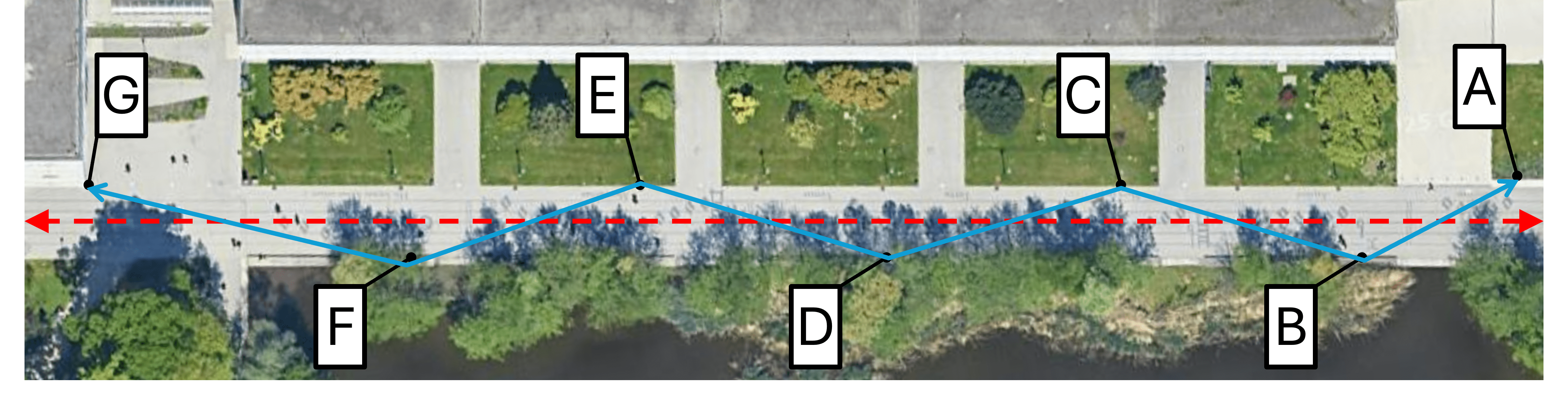}
    \caption{Experiment procedure: Participants walk between signs A–G, following a path similar to the blue route, while the vehicle moves along the red path.}
    \label{fig:experiment_path}
\end{figure}

\begin{figure}
    \centering
    \includegraphics[scale=0.5]{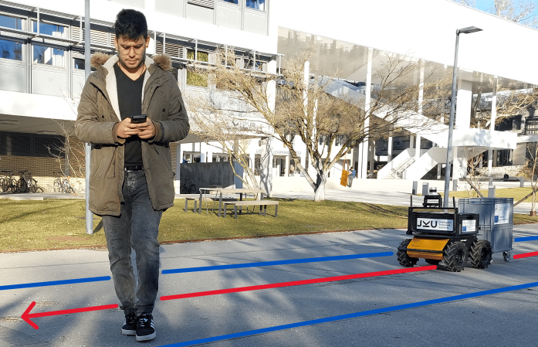}
    \caption{The figure shows a participant crossing the track of the vehicle. The participant is warned about the approaching vehicle when they enter the area between the two blue highlighted lines, parallel to the fixed track of the vehicle marked in red.}
    \label{fig:warning_position}
\end{figure}

The experiment was conducted on the campus of the Johannes Kepler University in Linz, Austria, with 10 participants in a pedestrian area. A remotely controlled robot mimicked the behavior of an AV (responding to dynamic obstacles, executing tasks without direct human presence, performing real-time decision-making and movement automation) by navigating a predefined route. To this end, a Clearpath' Husky 200 with a Zed2i camera for object detection, localization, and tracking was utilized. The camera detected the position and velocity of all pedestrians within its field of view.
Participants were instructed to walk between designated locations while the vehicle followed a fixed track, similar to a tram. Figure \ref{fig:experiment_path} illustrates the paths of both the participants and the vehicle. A close-up photo of the experiment is shown in Figure \ref{fig:warning_position}. The fixed track of the vehicle is highlighted in red, with its safety boundaries marked by blue lines. When a participant enters the safety boundaries in front of the vehicle, a warning is issued. Three types of warnings were tested: an audio signal emitted by the vehicle, a warning displayed on the trivia app, and both warnings presented simultaneously. The type of warning issued was randomized, additionally, a baseline condition was established by randomly repressing the warning signal to assess the participant's natural response, allowing us to measure the impact of different warning signals and the effect of warnings overall. Those cases are marked as "no warning" in the remainder of the study. Randomly turning off the warning helped stop participants from getting too used to it.

\begin{table}
\centering
\caption{List of the configurations and their abbreviations for the experiment. Each participant crossed 12 times with each distraction and experienced each type of warning twice. The abbreviations are generated by the type of distraction and warning, e.g. HP/NW defining the configuration where the participant was distracted by headphones, Trivia and received no warning}
\label{tab:configs}
\begin{tabular}{lll}
Configurations & Distraction          & Warning    \\ \hline
HP/NW          & Headphones \& Trivia & No warning \\
HP/R           & Headphones \& Trivia& Robot           \\
HP/A           & Headphones \& Trivia& App           \\
HP/R\&A        & Headphones \& Trivia& Robot \& App           \\
NHP/NW               &Trivia& No warning           \\
NHP/R &Trivia& Robot           \\
NHP/A               &Trivia& App           \\
NHP/R\&A               &Trivia& Robot \& App          
\end{tabular}
\end{table}

Accidents involving trams have increased due to pedestrians being visually distracted by mobile phone use and failing to hear the warning signal while wearing headphones. To replicate these conditions, participants in the experiment answered trivia questions on a smartphone while walking between the designated locations. Additionally, the experiment was repeated with the participant listening to music with headphones while answering trivia questions. The headphones served to further distract participants and to block any auditory cues from the vehicle.


During the experiment, participants walked twice from point A to point G and back, crossing paths with the vehicle twelve times in total. Table \ref{tab:configs} lists all tested interaction configurations and their corresponding abbreviations used throughout the paper.

\section{Data Processing}
\label{sec:Methodology}

Manual labeling was unnecessary for extracting participant trajectories, as the Zed2 camera provided their position and velocity.

For each participant the trajectories were classified into two categories
\begin{itemize}
    \item crossing: the participant crossed the vehicle's path in front of the vehicle
    \item stopping: the participant waited for the vehicle to pass before crossing behind it.
\end{itemize}
This classification was necessary to identify the most appropriate safety metrics for each scenario. 
Within each category, the trajectory was further classified depending on the type of warning the participant received, or whether there was no warning.
The independent variables, type of warning and baseline conditions, affect the dependent variables (Surrogate Safety Measures), which measure participant behavioral adjustments and preferred safety distance in this interaction. Observing these variations provides insights into how participants adapt their behavior in response to different warning signals.



\subsection{Surrogate safety measures}

Since the scenarios in this study closely resemble intersections, we selected two commonly used SSMs for such situations: Post Encroachment Time (PET) and Time-To-Collision (TTC). PET is particularly suited for scenarios where the vehicle and participant's path intersect, but don't collide.  TTC can capture safety information even when the pedestrian stops before crossing the vehicle's path, as it is mostly used as a predictive measure.

Note that PET was only calculated for interactions where the participant crossed the fixed track in front of the vehicle.

\subsubsection{Post Encroachment Time (PET)}
\begin{figure}
    \centering
    \includegraphics[scale=0.35]{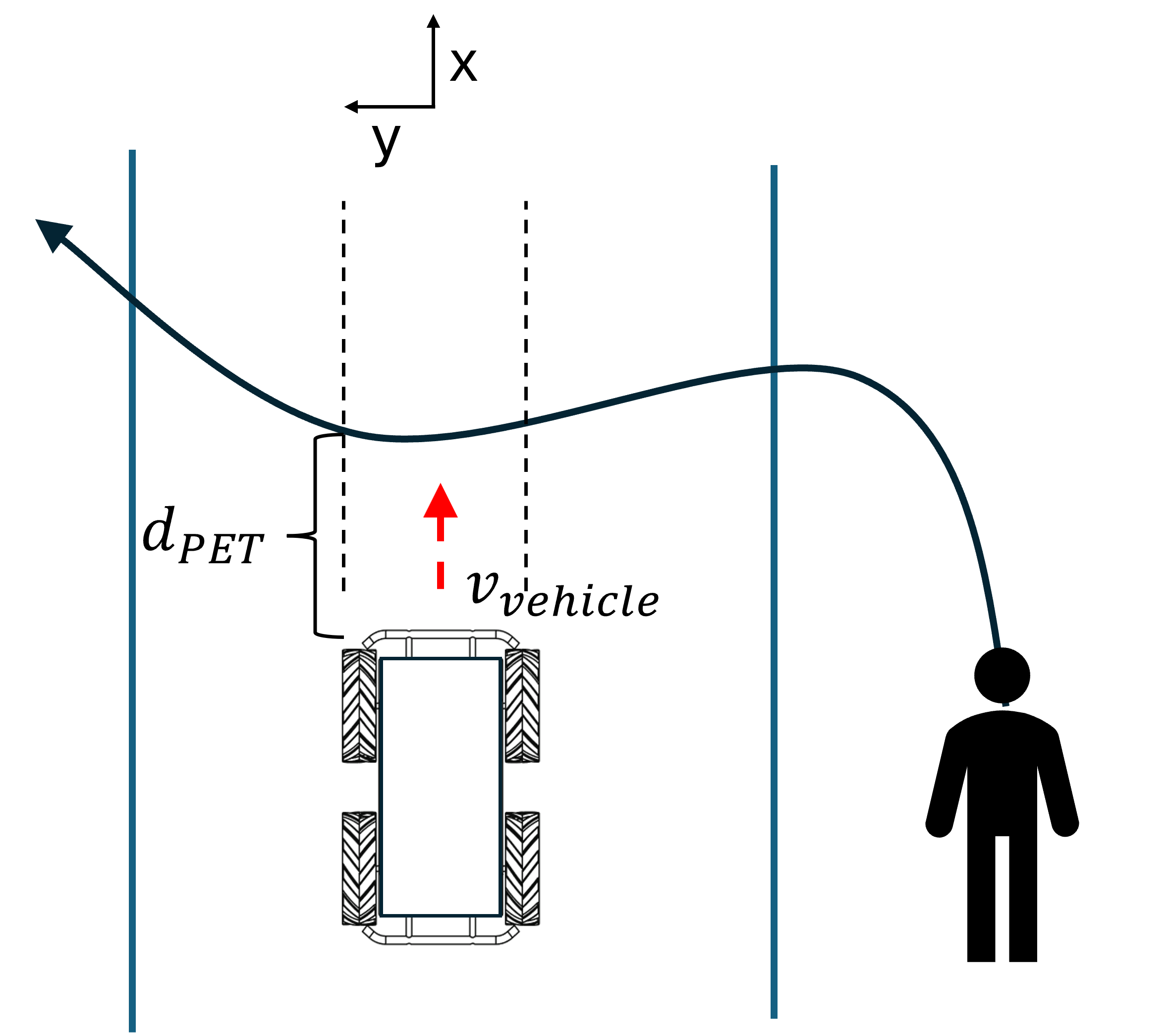}
    \caption{Visualization of the PET calculation for a crossing scenario. The participant moves along the black curve in the reference frame of the vehicle, and exits the vehicle's path at a relative distance $d_{PET}$. Dividing the distance by the speed $v_{vehicle}$ of the vehicle we obtain the PET.}
    \label{fig:PET_situation}
\end{figure}

PET is defined as the time interval between the departure of the first road user from a specific point of intersection and the subsequent arrival of the second road user at the same point. In this experiment, PET was calculated as the distance between the participant leaving the vehicle's path and the vehicle's location at that moment, divided by the vehicle's speed at that instance (see Figure \ref{fig:PET_situation} and Equation \ref{eq:PET}). Note that forward motion for the vehicle is defined by a positive speed in $x$ direction.

\begin{equation}
\label{eq:PET}
    PET = \frac{d_{PET}}{v_{vehicle,x}},
\end{equation}

\subsubsection{Time-To-Collision (TTC)}

TTC is defined as the time remaining until two road users, assuming constant speed and heading, would collide if no preventive actions were taken. Along the trajectories of the vehicle and pedestrians, a TTC value is calculated at each time step.

\begin{figure}
    \centering
    \includegraphics[scale=0.35]{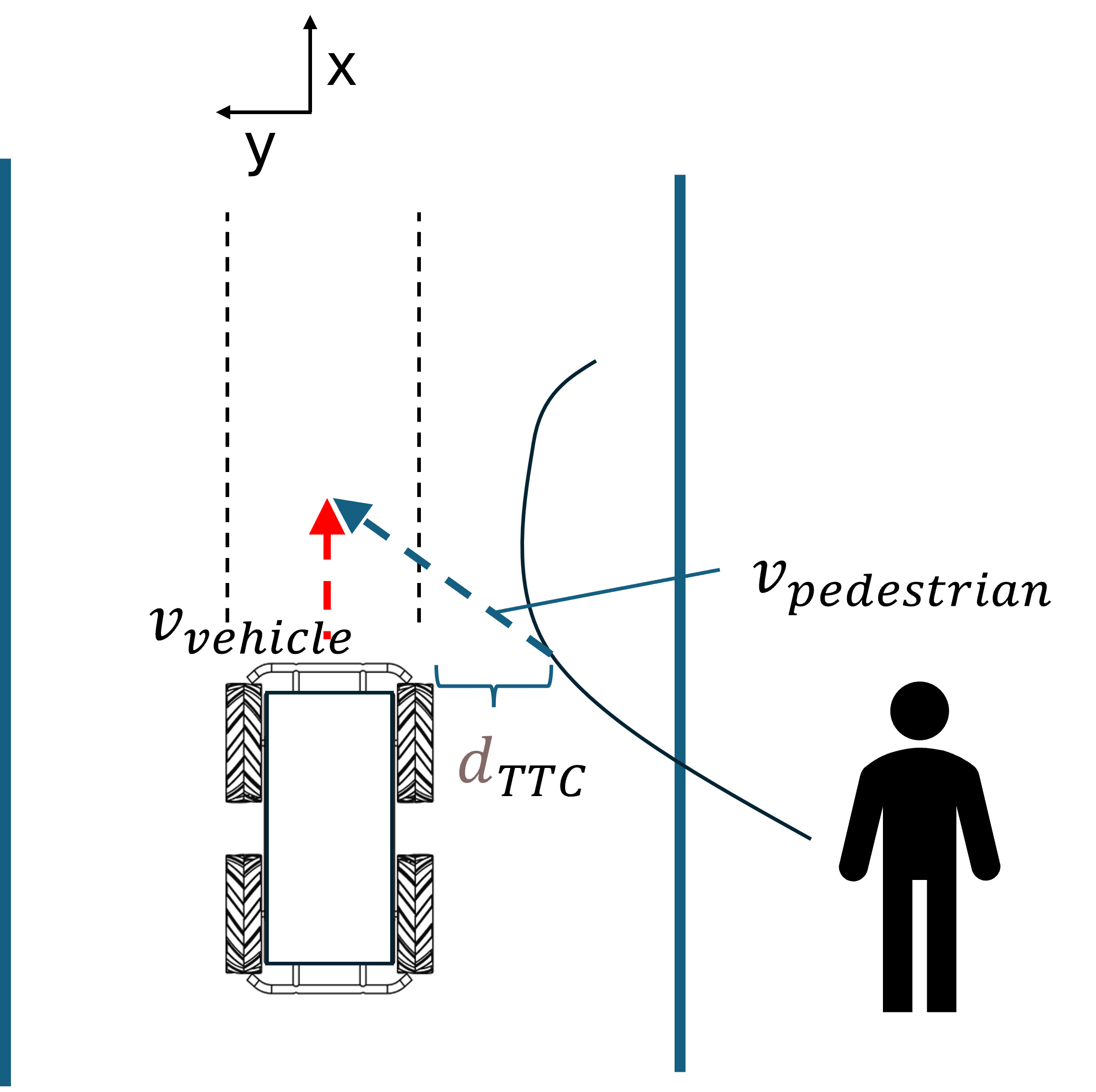}
    \caption{The potential collision between the pedestrian and the vehicle is predicted based on the pedestrian's continued movement with velocity $v_{pedestrian}$ and the vehicle's continued movement with velocity $v_{vehicle}$, reaching the point defined by the distance to the vehicle, $d_{TTC}$. This calculation is conducted at each timestep as the pedestrian progresses along the black curve. The minimum TTC value is considered to evaluate the safety of the interaction.}
    \label{fig:TTC_situation}
\end{figure}

TTC was computed for trajectories where the centerline remained uncrossed, meaning the pedestrian would cross after the vehicle passed. The interaction's safety was defined by the minimum observed TTC during the approach. In cases where no collision course existed, TTC was not calculable. The TTC calculation is depicted in Figure \ref{fig:TTC_situation} and equation \ref{eq:TTC}.

\begin{equation}
\label{eq:TTC}
    TTC = \frac{d_{TTC}}{-v_\mathrm{pedestrian,y}},
\end{equation}
where $v_\mathrm{pedestrian,y}$ represents the pedestrian's speed in the $\mathrm{y}$ (lateral) direction, and $d_{TTC}$ denotes the distance in the y direction.

\subsection{Statistical analysis}

By calculating the safety metrics (PET and TTC) for all interactions between pedestrians and vehicle, we obtained two distributions for each configuration in Table \ref{tab:configs}. We define the baseline no-warning distributions from the aggregated distributions of configuration HP/NW and NHP/NW. We then compared these baseline distributions with those from interactions in which warnings were issued.

Since the same subjects perform the different conditions, the data points are paired, meaning each subject serves as their own control. Yet, the different configurations from Table \ref{tab:configs} are distributed differently in the two interaction categories, crossing and stopping, for each participant. Thus it is necessary to first aggregate the average value for a specific configuration for each participant in order to compare the distribution of PET or TTC for different configurations.

As the distributions are not normally distributed (according to a Shapiro–Wilk test), a non-parametric statistical test, the Wilcoxon signed-rank test, is utilized for the comparison. It evaluates the null hypothesis that the distribution of differences between paired samples is centered at zero. With a p-value less than 0.05 the the null hypothesis is rejected in favor of the alternative, that the distribution is centered above zero.

\section{Results}
\label{sec:results}

The experiment was conducted with 10 participants,  resulting in a total of 224 vehicle-pedestrian interactions.
However, only 187 interactions provided complete trajectories of the participants, mainly due to interference from other pedestrians. Additionally, three interactions in which pedestrians and other obstacles forced the vehicle to stop were excluded from further analysis.
Qualitative observations revealed substantial variability in participant responses, ranging from consistently stopping to yield to the vehicle on the fixed track, to disregarding its path regardless of the warning.

Table \ref{tab:trajectories} lists the number of interactions per type, further divided into warning/no warning. Note that the HP/R configuration, in which participants wore headphones and received warnings only from the vehicle, was not included in the warning/no-warning categories and is not included in further analysis unless explicitly mentioned. This exclusion was due to ambiguity regarding the clarity of the warning, as participants later reported difficulties and inconsistency in perceiving it.

\subsection{TTC}
\begin{figure}
    \centering
    \includegraphics[scale=0.6]{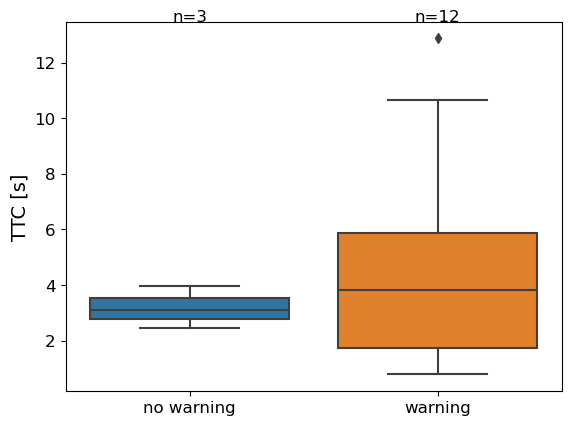}
    \caption{Visualization of the distributions of TTC values for all stopping scenarios, split into warning and no warning interactions. The center line of each box indicates the mean values. The boxed area contains 50\% of the points. Above each configuration is the number of interactions of that specific configuration. Note that in three stopping interactions with a warning no TTC could be determined due to the absence of a collision path.}
    \label{fig:TTC_overall}
\end{figure}

As the number of interactions where the participant stopped and yielded to the vehicle was too low for statistical testing, the results of the distribution of TTC are only visualized in Figure \ref{fig:TTC_overall}. The figure shows the distribution of the minimal TTC calculated during the approach of the participant to the vehicle's fixed track. The primary difference between the two distributions is the frequency of stopping events: only 3\% of interactions without a warning led to the participant stopping and yielding, whereas this increased to 18\% when a warning was provided.

\begin{table}
\centering
\caption{Number of interactions per type and warning category. Interactions in the HP/R configuration are excluded when distinguishing between warning and no-warning interactions, resulting in totals that do not fully sum up.}
\label{tab:trajectories}
\begin{tabular}{llll}
Type of interaction & warning/no warning & \multicolumn{2}{l}{\# of interactions} \\ \hline
crossing            & warning            & 67        & \multirow{2}{*}{164}       \\ \cline{1-3}
crossing            & no warning         & 84        &                            \\ \hline
stopping            & warning            & 15        & \multirow{2}{*}{20}        \\ \cline{1-3}
stopping            & no warning         & 3        &                           
\end{tabular}
\end{table}

\subsection{PET}
\begin{figure}
    \centering
    \includegraphics[scale=0.5]{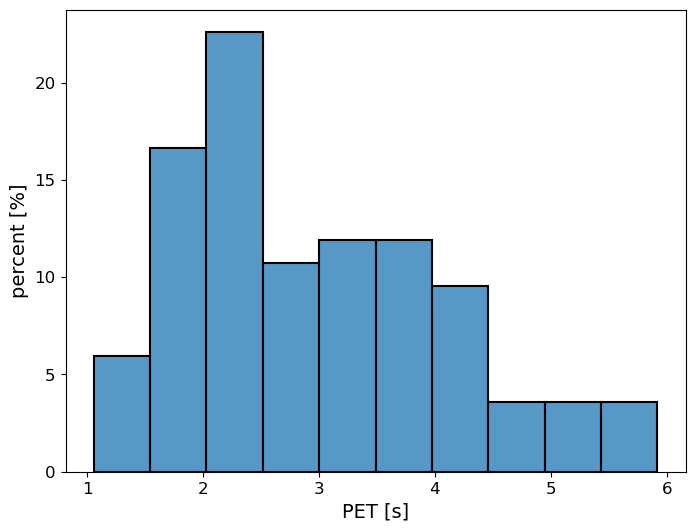}
    \caption{Histogram of PET values for all crossing scenarios without issued warnings.}
    \label{fig:PET_no_warning}
\end{figure}

\begin{figure}
    \centering
    \includegraphics[scale=0.5]{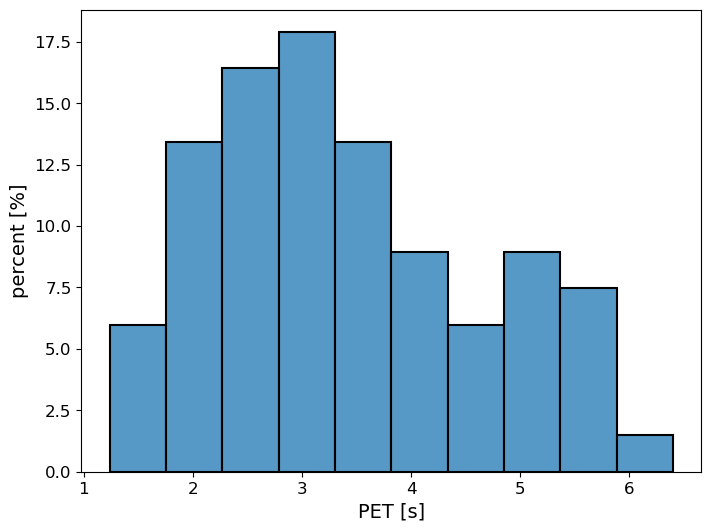}
    \caption{Histogram of PET values for all crossing scenarios in which a warning was issued.}
    \label{fig:PEt_warning}
\end{figure}

Figure \ref{fig:PET_no_warning} presents the distribution of PET values for interactions where participants received no warnings, with a mean of 2.97 and a standard deviation of 1.14. The distribution of PET for interactions where participants received warnings is shown in Figure \ref{fig:PEt_warning}, with a mean of 3.41 and a standard deviation of 1.26. 
However, the Wilcoxon signed rank test does not reject the Null hypothesis as defined in Section \ref{sec:Methodology} with $p_{value} = 0.10 > 0.05$.

\begin{figure}
    \centering
    \includegraphics[scale=0.5]{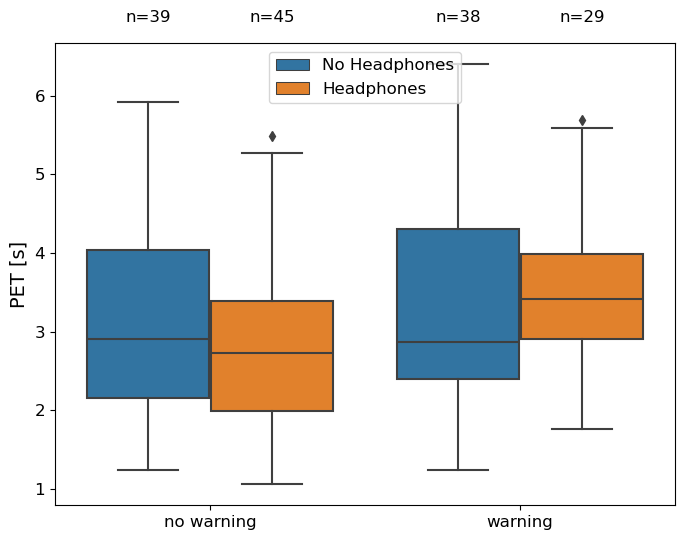}
    \caption{Boxplot of the distributions of PET values for all crossing scenarios split by warning/no warning and type of distraction. The center line of each box indicates the mean values. The boxed area contains 50\% of the points. Above each configuration is the number of interactions of that specific configuration.}
    \label{fig:PET_boxplot_mode}
\end{figure}

Dividing the previously mentioned distributions of PET by distraction type, i.e. with headphones or without headphones, yields the Figure \ref{fig:PET_boxplot_mode}.
In interactions where the participants were wearing headphones, the distribution of PET values is statistically significant higher when a warning was issued compared to when it was not (Wilcoxon signed rank test $p_{value} = 0.01$).

\begin{figure}
    \centering
    \includegraphics[scale=0.6]{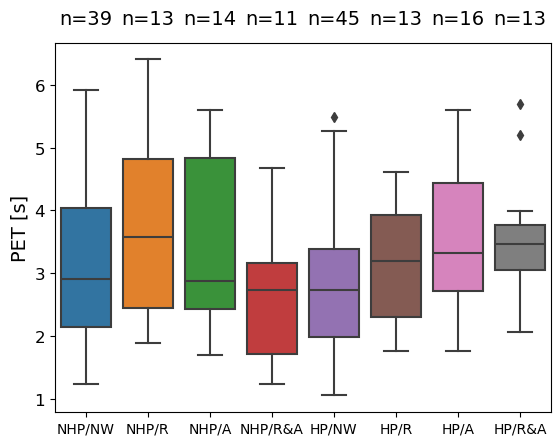}
    \caption{Visualization of the distributions of PET values split by configuration. The center line of each box indicates the mean values. The boxed area contains 50\% of the points. Above each configuration is the number of interactions of that specific configuration.}
    \label{fig:PET_boxplots}
\end{figure}

Further splitting the distributions according to the configurations defined in Table \ref{tab:configs} results in the distributions presented in Figure \ref{fig:PET_boxplots}.
Table \ref{tab:Wilcoxon_all} presents the pairwise Wilxocon signed rank test of those distributions. The test assumes an alternative hypothesis where the distribution in each row is shifted toward higher values compared to the corresponding column distribution. Most comparisons were not significant ($p_ {value} > 0.05$), however, both HP/A and HP/R\&A showed a statistically significant increase compared to HP/NW.
Therefore, for interactions with the participant wearing headphones, PET appears to capture the participants reactions to the vehicle moving along a fixed track, yet the reaction does not depend on the type of warning.
Notably, the vehicular robot produced significant noise while in motion, potentially allowing participants without headphones to estimate the vehicle's proximity despite the absence of an explicit warning.

In addition, in Table \ref{tab:Wilcoxon_all} NHP/R shows a statistically significant increase compared to HP/NW and NHP/A. A possible reasoning for this is that a warning through only the smartphone is perceived as less urgent than a warning through the vehicle's horn.

\begin{table}
\caption{Results of the pairwise Wilcoxon signed rank test for the different configurations of vehicle-pedestrian interaction, as defined in Table \ref{tab:configs}. An entry of \textbf{s}$^*$ signifies that the Null hypothesis was rejected with a $p_ {value} < 0.05$ and the PET distribution from the configuration defined by the row is shifted positively compared to the one defined by the column. An entry of n signifies that the Null hypothesis was not rejected.}
\centering
\setlength\tabcolsep{6pt}
\begin{tabular}{l|llllllll}
         & \rotatebox{90}{HP/NW} & \rotatebox{90}{HP/R} & \rotatebox{90}{HP/A} & \rotatebox{90}{HP/R\&A} & \rotatebox{90}{NHP/NW} & \rotatebox{90}{NHP/R} & \rotatebox{90}{NHP/A} & \rotatebox{90}{NHP/R\&A} \\
        \hline
        HP/NW   & -   & n   & n   & n   & n   & n   & n   & n   \\ \hline
        HP/R    & n   & -   & n   & n   & n   & n   & n   & n   \\ \hline
        HP/A    & \textbf{s}$^*$   & n   & -   & n   & n   & n   & n   & n  \\ \hline
        HP/R\&A & \textbf{s}$^*$   & n   & n   & -   & n   & n   & n   & n   \\ \hline
        NHP/NW  & n   & n   & n   & n   & -   & n   & n   & n  \\ \hline
        NHP/R   & \textbf{s}$^*$   & n   & n   & n   & n   & -   & \textbf{s}$^*$   & n   \\ \hline
        NHP/A   & n   & n   & n   & n   & n   & n   & -   & n   \\ \hline
        NHP/R\&A & n   & n   & n   & n   & n   & n   & n   & -   \\ 
        \hline
\end{tabular}
\label{tab:Wilcoxon_all}
\end{table}

\section{Conclusion and Future Work}
\label{sec:Conclusion}

An experiment was conducted to assess traffic safety in a shared space with trams and pedestrians using two Surrogate Safety Measures, Post-Encroachment-Time (PET) and Time-To-Collision (TTC).
Warnings before crossing the vehicle's path led to a statistically significant higher safety distance for interactions where the pedestrian, utilizing headphones, crossed the path of the AV in front of it, independent on the type of warning. This result aligns with previous findings in shared spaces (\cite{Morales_2020} and \cite{Hussein_2016}), indicating that safety outcomes are not dependent on warnings. Results on the effect of warnings when the pedestrian was not wearing headphones were still inconclusive.
Pedestrians were also observed to stop and yield to the AV more frequently in interactions with a warning. However, analyzing TTC values for this type of interaction was not feasible due to the small sample size, highlighting the need for further studies.



\section*{ACKNOWLEDGMENT}

The work has been conducted as a part of OptiPEx project (No.101146513) funded by the European Union. Views and opinions expressed are however those of the author(s) only and do not necessarily reflect those of the European Union or CINEA. Neither the European Union nor the granting authority can be held responsible for them.


\bibliographystyle{IEEEtran}
\bibliography{bliblio}

\end{document}